\pdfoutput=1

\documentclass[11pt]{article}

\usepackage[preprint]{acl}

\usepackage{subfiles}
\usepackage{times}
\usepackage{latexsym}
\usepackage{booktabs}
\usepackage{arydshln}
\usepackage{float}
\usepackage{subcaption} 
\usepackage[normalem]{ulem}

\usepackage[T1]{fontenc}

\usepackage[utf8]{inputenc}

\usepackage{microtype}

\usepackage{inconsolata}

\usepackage{graphicx}
\usepackage{enumitem}
\usepackage{adjustbox}
\setlist[itemize]{noitemsep, topsep=0pt, leftmargin=11pt}
\setlist[enumerate]{noitemsep, topsep=0pt, leftmargin=11pt}

\newcommand{\divider}
{\par\bigskip\hrule\bigskip\par}
\title{Which LLMs Get the Joke? Probing Non-STEM Reasoning Abilities with HumorBench}

\author{
 \textbf{Reuben Narad\textsuperscript{1}},
 \textbf{Siddharth Suresh\textsuperscript{2}},
 \textbf{Jiayi Chen\textsuperscript{2}},
 \textbf{Pine S.L. Dysart-Bricken\textsuperscript{1}},
\\
 \textbf{Bob Mankoff\textsuperscript{3}},
 \textbf{Robert Nowak\textsuperscript{2}},
 \textbf{Jifan Zhang\textsuperscript{2}},
 \textbf{Lalit Jain \textsuperscript{1}},
\\
 \textsuperscript{1}University of Washington, Seattle,
 \textsuperscript{2}University of Wisconsin-Madison,\\
 \textsuperscript{3}Air Mail and Cartoon Collections\\
 \small{
   \textbf{Correspondence:} \href{mailto:rnarad@uw.edu}{rnarad@uw.edu}, \href{mailto:jifan@cs.wisc.edu}{jifan@cs.wisc.edu}
 }
}

\begin{document}
\maketitle
\begin{abstract}
We present HumorBench, a benchmark designed to evaluate large language models' (LLMs) ability to reason about and explain sophisticated humor in cartoon captions. As reasoning models increasingly saturate existing benchmarks in mathematics and science, novel and challenging evaluations of model intelligence beyond STEM domains are essential.
Reasoning is fundamentally involved in text-based humor comprehension, requiring the identification of connections between concepts in cartoons/captions and external cultural references, wordplays, and other mechanisms.
HumorBench includes approximately 300 unique cartoon-caption pairs from the New Yorker Caption Contest and Cartoonstock.com, with expert-annotated evaluation rubrics identifying essential joke elements. LLMs are evaluated based on their explanations towards the humor and abilities in identifying the joke elements. To perform well on this task, models must form and test hypotheses about associations between concepts, potentially backtracking from initial interpretations to arrive at the most plausible explanation.
Our extensive benchmarking of current SOTA models reveals three key insights: (1) LLM progress on STEM reasoning transfers effectively to humor comprehension; (2) models trained exclusively on STEM reasoning data still perform well on HumorBench, demonstrating strong transferability of reasoning abilities; and (3) test-time scaling by increasing thinking token budgets yields mixed results across different models in humor reasoning.

\end{abstract}

\section{Introduction}
Recent advances in large language models and reasoning techniques have led to the saturation of many existing benchmarks, particularly in STEM domains such as mathematics and programming, where frontier models now approach or exceed human-level performance~\citep{phi4reasoning2025, olymmath2025, quan2025codeelo}. This progression highlights the need for novel and challenging evaluations that can meaningfully differentiate model capabilities and provide insights into their reasoning processes. Non-STEM reasoning tasks, particularly those involving cultural understanding and implicit knowledge, represent underexplored territories for model evaluation.

Humor comprehension represents a particularly challenging frontier for artificial intelligence~\citep{hessel-etal-2023-androids, zhang2024humor, zhou2025bridging, kazemi2025big, liang2025yesmeetsbutlarge}. Although large language models (LLMs) excel across many domains, understanding humor still requires sophisticated reasoning that integrates context, cultural knowledge, and implicit connections. These challenges make humor an ideal testbed for evaluating advanced reasoning in AI systems.

\begin{figure*}[]
    \centering
    \includegraphics[width=\linewidth]{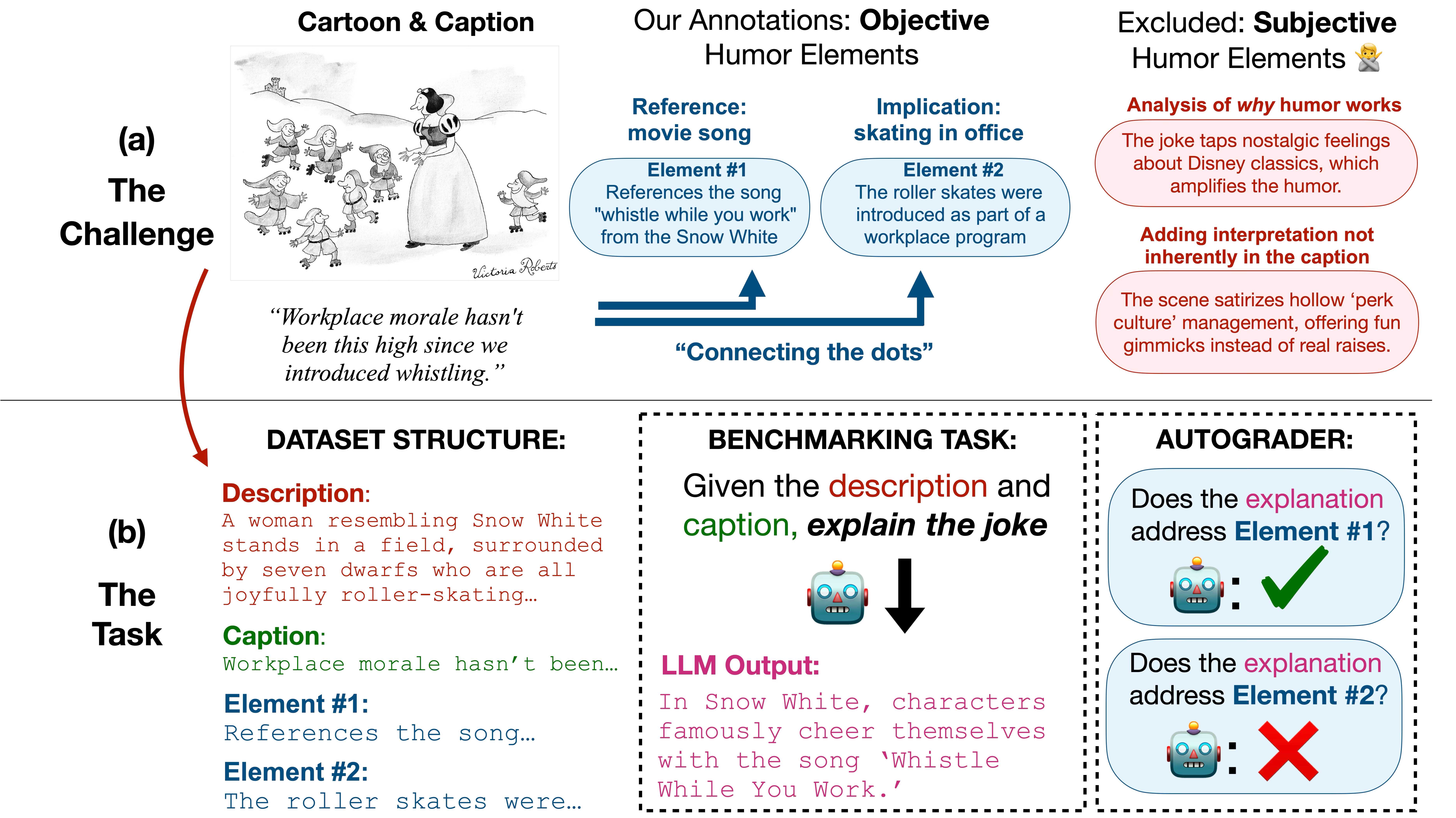}
    \caption{Overview of our humor analysis approach. (a) We distinguish between \textbf{objective} and \textbf{subjective} components of a joke. To convert the open-ended task of humor explanation into a fair benchmark, we focus exclusively on \textbf{objective} elements. (b) Overview of the dataset, benchmark task, and grading scheme in HumorBench. Each cartoon-caption pair contains one or more ``element'' annotation. For the benchmark, an LLM is tasked with explaining the joke in the caption. An autograder evaluates if the explanation contains each element.}
    \label{fig:intro}
    \vspace{-\intextsep}
\end{figure*}



We present \textbf{HumorBench}, a benchmark that evaluates LLMs' ability to explain sophisticated cartoon-caption humor by identifying the mental leaps connecting visuals, captions, and external knowledge (Figure~\ref{fig:intro}). For each pair, we annotate the objective elements essential for comprehension, creating a ground truth focused on factual connections rather than subjective appreciation.

We benchmark both a standard set and a harder subset. On HumorBench-hard, which features more complex examples requiring multiple reasoning steps or obscure cultural knowledge, no current LLM exceeds 60\% accuracy.
Our benchmarking of current state-of-the-art models reveals two key findings:
\begin{enumerate}
    \item We observe a high correlation between performance on HumorBench and existing STEM benchmarks, suggesting a significant transfer of general reasoning abilities to humor comprehension tasks.

    \item Even models trained exclusively on STEM reasoning tasks (e.g., mathematical problem-solving) perform well on HumorBench, indicating that abstract reasoning skills acquired in one domain can transfer effectively to humor comprehension.

    \item Test-time scaling measures for humor reasoning yield mixed results, indicating that simply increasing computational resources at inference time does not consistently improve performance on this challenging domain.
\end{enumerate}

\subsection{Why Another LLM Humor Benchmark?} \label{ssec:novelty}
Several benchmarks have already focused on measuring LLMs' capabilities around humor. Specifically, \citet{hessel-etal-2023-androids} and \citet{zhang2024humor} both build upon the New Yorker Caption Contest (NYCC) dataset—a weekly feature by the New Yorker magazine where readers submit funny captions for cartoons (see Figure~\ref{fig:intro} for an example).
\citet{hessel-etal-2023-androids} created three benchmarks using this dataset: ranking the funniness of caption pairs, matching cartoons to valid captions, and explaining the humor behind captions. However, these benchmarks simultaneously measure two distinct capabilities: (1) understanding the intended jokes (objective elements) and (2) aligning with individual and subgroup humor preferences (subjective factors). As \citet{zhou2025bridging} points out, performance on these previous benchmarks is heavily influenced by an LLM's ability to align with specific audience preferences rather than directly measuring its reasoning about the jokes themselves. 

Our benchmark, HumorBench, addresses this limitation by focusing solely on the objective elements of humor comprehension, specifically measuring the \emph{humor reasoning abilities} required to understand cartoons and their captions. 
As validated by our experimental findings, LLMs' performance on HumorBench correlates well with their performance on other reasoning benchmarks.

\section{Related Work} \label{sec:related}

\noindent\textbf{Reasoning‑focused language models.}  
Recent advances in large language models (LLMs) have seen the emergence of specialized reasoning models that excel at logical deduction, mathematical problem-solving, and multi-step reasoning while maintaining strong general language capabilities. These reasoning-enhanced models employ various approaches: training-focused methods like those used by \emph{Minerva} \citep{Lewkowycz2022Minerva}, \emph{WizardMath} \citep{Luo2023WizardMath}, and \emph{Phi-4-Reasoning} \citep{Abdin2025Phi4} leverage carefully curated STEM-heavy corpora; inference-time techniques boost reasoning without changing model weights, including \emph{Self-consistency} \citep{Wang2023SelfConsistency}, \emph{Tree-of-Thought} methods in \emph{DeepSeek-Math} \citep{Shao2024DeepSeekMath}, \emph{Least-to-most} prompting \citep{Zhou2023LeastToMost}, and \emph{Process supervision} \citep{Lightman2023ProcessSupervision}; while hybrid approaches like \emph{MAmmoTH} \citep{Yue2023MAmmoTH} combine diverse training data with structured inference protocols, \emph{ToRA} \citep{Gou2023ToRA} integrates formal verification systems, and \emph{MathGLM} \citep{Yang2024MathGLM} combines symbolic computation with natural language reasoning. Models like \emph{Gemini Ultra} \citep{Google2024Gemini} and \emph{Claude 3.5} \citep{Anthropic2024Claude} achieve strong reasoning through both architectural innovations and sophisticated training, suggesting that advances in machine reasoning now follow multiple complementary paths rather than relying solely on parameter count \citep{Wei2024EmergentReasoning}.

\noindent\textbf{Humour benchmarks.}  
Beyond simple joke generation, several resources now probe LLM humour competence. \citet{hessel-etal-2023-androids} introduces three New Yorker cartoon-caption subtasks that test multimodal humour understanding and explanation. For word-play, the \emph{ExPUNations} corpus augments classic pun datasets with human-written explanations and funniness ratings \citep{Sun2022ExPun}, while  \citet{Xu2024GoodPun} systematically benchmarks pun recognition, explanation and creation. Complementing these datasets, \citep{Ermakova2025JOKER} Lab provides reusable test collections for humour-aware information retrieval. 

\noindent\textbf{Open‑ended evaluation frameworks.}  
Automatic grading of creative, unconstrained outputs increasingly relies on the \emph{LLM-as-Judge} paradigm. \textsc{G-Eval} couples chain-of-thought GPT-4 judging with a form-filling rubric, achieving human-level reliability on summarisation and dialogue \citep{Liu2023GEval}. \textsc{MT-Bench} and its crowdsourced \emph{Chatbot Arena} show that GPT-4 judges agree with human preferences on multi-turn instruction following in $\sim$80\% of cases \citep{Zheng2023MTBench}. Going further, \textsc{PaperBench} grades agents on reproducing ICML-level research papers with hierarchical GPT-4 rubrics and expert audits \citep{Starace2025PaperBench}. We adopt a similar rubric-guided judging scheme but focus specifically on humour reasoning, enabling systematic comparison of explanation quality across models.

\section{HumorBench}
\subsection{Main Benchmark Task}
HumorBench frames humor understanding as an open-ended task: given a textual description of a cartoon and its caption, a model must articulate in its own words the underlying joke. We deliberately avoid the multiple-choice or ranking formats common in existing humor benchmarks because, for creative tasks, fixed answer sets can (i) inadvertently hint at the punchline and (ii) fail to accommodate the diverse range of valid explanations a competent reader might produce.

To make this free-form setting automatically gradable, we distill each cartoon into a concise rubric of 1–3 objective ``elements." An element represents a single, easily verifiable fact that any correct explanation must include (e.g., in NYCC Contest \#665, the observation that ``the shark interprets the swimmer as groceries"), as shown in Figure~\ref{fig:hardset}. This approach allows for creative expression while maintaining consistent evaluation standards (see Appendix~\ref{sec:Explainer_Prompt} for the complete prompt).

\subsection{Dataset: Cartoon and Caption Sources}
Our dataset comprises cartoons and captions from two primary sources: the New Yorker Caption Contest (NYCC) and Cartoonstock.com. We sourced NYCC captions from publicly available datasets~\citep{hessel-etal-2023-androids,jain2020new,zhang2024humor}, selecting only those ranked among the top 3 finalists to ensure each cartoon features a coherent, high-quality joke. For Cartoonstock cartoons, we utilized their original accompanying captions. Both sources specialize in dry, witty humor that demands sophisticated reasoning—often requiring multiple mental leaps to fully comprehend, as illustrated in Figure~\ref{fig:intro}.

While cartoons inherently include visual elements, our benchmark focuses on testing humor comprehension rather than visual interpretation capabilities. Therefore, we created detailed textual descriptions of each cartoon, carefully capturing all information necessary to understand the caption while maintaining neutrality. These descriptions include essential details about the setting, characters, visible emotions, and speaker identification, while deliberately omitting artistic style unless directly relevant to the joke. For researchers interested in extending this to a multimodal benchmark, we provide source links to the original images—NYCC images are available through~\citep{hessel-etal-2023-androids,jain2020new}, while Cartoonstock images require licensing.

\subsection{Dataset: Element Annotation}
The core labels in our dataset are the \emph{element} annotations assigned to each cartoon–caption pair. For every pair, we hand-annotated one to three elements—concise, direct statements that capture the objective components essential to understanding the joke.
As discussed in Section~\ref{ssec:novelty}, comprehending cartoon humor requires two distinct capabilities: understanding the objective content of the joke and recognizing the subjective aspects that influence audience reception. Figure~\ref{fig:intro} illustrates this distinction—subjective explanations focus on audience reactions (which vary between individuals), while objective elements center on content comprehension. Our benchmark specifically targets these objective elements, which require identifying the mental leaps necessary to "get" the joke through recognizing references, wordplay, implications, or similar mechanisms.
The autograder then evaluates LLM explanations against these elements, verifying that each explanation adequately covers the fundamental objective components of the humor, ensuring a fair and consistent assessment across different models.

In summary, to make this task easily gradable, annotations follow a set of deliberate guidelines:
(1) Elements must be short, direct, and easily verifiable from the description and caption;
(2) An element addresses exactly one concept. Bundling multiple ideas may add noise by forcing the autograder to guess about partial correctness;
(3) Elements deliberately avoid adding bias from subjective opinion about humor.


\begin{figure*}[th]
  \centering
  \setlength{\tabcolsep}{0pt}

  \begin{tabular}{@{}p{0.40\textwidth}@{\hspace{1.2em}}p{0.55\textwidth}@{}}
    \begin{minipage}[th]{\linewidth}\vspace{0pt}
      \includegraphics[width=\linewidth]{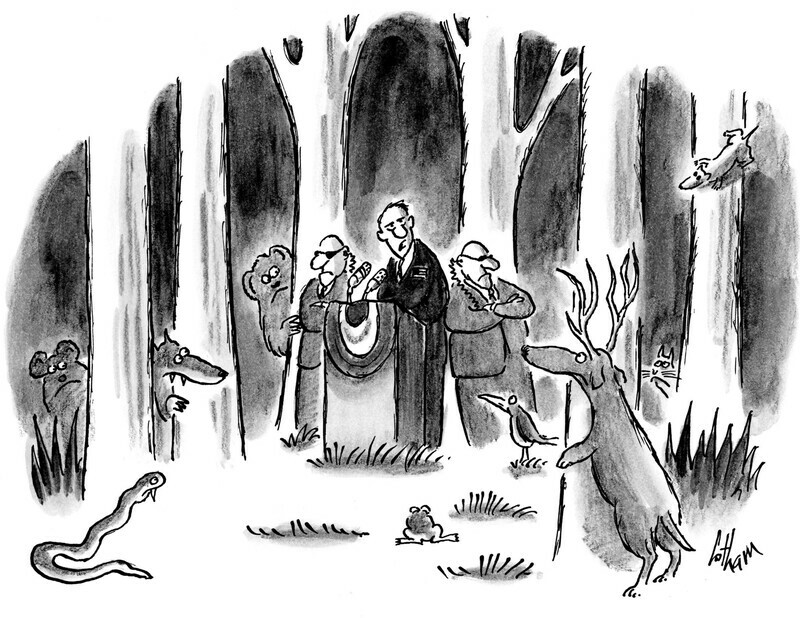}
    \end{minipage}
    &
    \begin{minipage}[th]{\linewidth}\vspace{0pt}
      \normalsize
      \raggedright
      \textbf{NYCC Contest 167}\\

      \textbf{Description:} In a forest clearing with tall tree trunks,  
      a man in a suit with an American-flag lapel pin speaks at a podium  
      while aides stand behind him. Woodland animals—a deer, snake, frog,  
      and birds—peek from the trees and grass, watching.\\[0.8em]

      \textbf{Caption:} “As a weasel, I need your vote.”%

      \divider  
      \textbf{ELEMENT 1:} References the cliché insult of calling  
      politicians \emph{weasels}.\\[0.5em]
      \textbf{ELEMENT 2:} Plays on the dual meaning of “weasel” (literal  
      animal \& political pejorative), creating a pun.
    \end{minipage}
  \end{tabular}

  \caption{Example \textsc{HumorBench} annotation. The cartoon
           (\textit{left}) is paired with its description, caption, and
           two hand-labeled joke elements (\textit{right}).}
  \label{fig:humorbench-example}
\vspace{-\intextsep}
\end{figure*}

\subsection{Dataset Refinement}
For an LLM evaluation to provide trustworthy results, the underlying dataset must be both accurate and internally consistent. We initially collected 655 unique element annotations, but despite careful guidelines, some entries proved vague or imprecise. To systematically improve quality, we implemented an iterative refinement process.

First, we generated sample explanations for each cartoon–caption pair, alternating randomly between GPT-4o and Claude 3.7 Sonnet. Each explanation was evaluated ten times by our autograder, with elements showing verdict disagreement exceeding 30\% flagged for review. These problematic cases were either refined or removed entirely. We repeated this quality control cycle until fewer than 5\% of annotations triggered inconsistency flags, ultimately resulting in 499 high-quality unique element annotations forming the foundation of HumorBench.

As an additional validation step, we invited a former chief cartoon editor of the New Yorker to review a random subset of 30 annotations. The editor confirmed that all elements were fair and accurately captured the essential components of each joke. Together, these atomic, objectively verifiable criteria create a robust rubric that enables our autograder to provide consistent and reliable evaluation at scale.

\subsection{Autograder and Evaluation}
During evaluation, an LLM judge assesses each model's explanation against individual elements to determine whether they adequately cover the essential components of the joke. This approach allows us to efficiently evaluate open-ended text generation at scale.

However, ensuring autograder consistency presents challenges, particularly for tasks that are inherently difficult for LLMs to understand~\citep{Min2020AmbigQA,Starace2025PaperBench}. To address this, we created a separate benchmark of 300 human expert judgments on explanations from three distinct LLMs: GPT-4o, Gemini 2.5 Pro, and Claude 3.7 Sonnet. Using GPT-4o as the autograder, we achieved 92\% accuracy overall:
\begin{table}[H]
\centering
\footnotesize
\setlength{\tabcolsep}{5pt}
\captionsetup[subfigure]{%
  labelformat=empty,        
  justification=raggedright,
  singlelinecheck=false,    
  skip=2pt                  
}
\begin{adjustbox}{max width=\linewidth}
\begin{tabular}{lccc}
\toprule
\textbf{Explainer Model} & \textbf{Acc. (\%)} & \textbf{FPR (\%)} & \textbf{FNR (\%)} \\
\midrule
\textit{Overall} (n=300) & 92.00 & 14.79 & 6.51 \\
\midrule
Gemini 2.5 Pro           & 93.00 & 10.00 & 6.25 \\
GPT-4o                   & 92.00 & 14.81 & 5.48 \\
Claude 3.7 Sonnet        & 91.00 & 19.57 & 7.80 \\
\bottomrule
\end{tabular}
\end{adjustbox}
\caption{Autograder performance (GPT-4o judge) on 300 human-labeled explanations.}
\label{tab:autograder-results}
\vspace{-\intextsep}
\end{table}
This validation provides two key insights. First, across all models, the autograder's false positive rate (FPR) substantially exceeded its false negative rate (FNR), indicating a leniency bias. This suggests that \emph{HumorBench scores should be interpreted as an upper bound on model performance}. Second, despite using GPT-4o as the autograder, we observed no significant advantage for GPT-4o-generated explanations compared to those from other models. Together, these findings confirm that our autograder provides a valid, albeit slightly optimistic, mechanism for large-scale evaluation.

\noindent\textbf{Length Control. } While models were instructed to keep responses under 200 words, some models exceeded this limit, particularly when reasoning traces were included in the final output. To ensure fair comparison, we truncated all model outputs to the last 1000 tokens.

\section{Experiments}
Along with creating the HumorBench evaluation, we extensively benchmarked current frontier models. For consistency, all models are given the same prompt and scaffolding describing the task (see Appendix ~\ref{sec:Explainer_Prompt}). 
We arrived at this prompt after validating across several different LLMs (Claude 3.7 Sonnet \cite{anthropic2025claude37sonnet}, GPT-4o \cite{openai2024gpt4o}, Gemini 2.5 Pro \cite{google2025gemini25pro}). While many LLMs had different API endpoints, we tried to maintain consistent parameters where possible. For example, all models had temperature set to 1 and external tool calling deactivated. Note, for all evaluations, autograders, and benchmarks, ``GPT-4o" refers to the gpt-4o-2024-08-06 release.


\subsection{Main Results}
\begin{figure}[th]
  \centering
  \includegraphics[width=\linewidth]{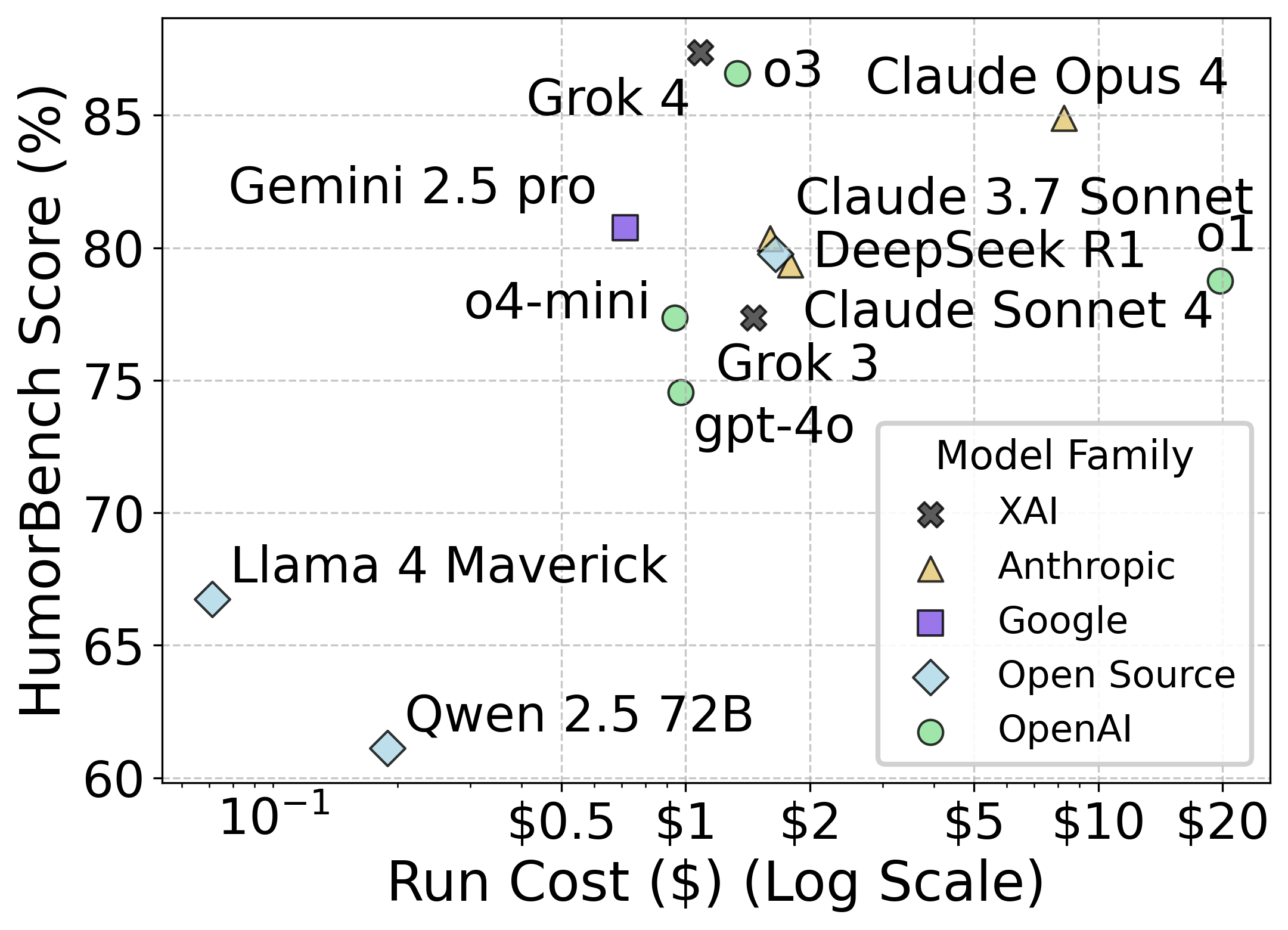}
  \caption{Benchmarking results on frontier models}
  \label{fig:main_bench}
\vspace{-\intextsep}
\end{figure}

In general, the results from the main benchmarking effort were unsurprising. As shown in Figure ~\ref{fig:main_bench}, OpenAI o3 \cite{openai2025o3} leads the pack at 87.5\% accuracy, dramatically ahead of other SOTA models (Gemini 2.5 Pro, Claude 3.7 Sonnet, and Deepseek R1 \cite{deepseek2025r1}), all achieving approximately 80\%.
In general, smaller models (like Llama 4 Maverick \cite{meta2025llama4maverick}, Qwen 2.5 \cite{alibaba2025qwen25}, and o3-mini \cite{openai2025o3mini}) performed worse.
We also found that newer versions of models generally dominate older versions of the same model, with o3 outperforming o1 \cite{openai2024o1} and Gemini 2.5 pro outperforming Gemini 1.5 pro.
In general, "reasoning" versions of models seemed to outperform the base versions of the same model. For example, DeepSeek R1 (79.8\%) strongly outperformed Deepseek V3 \cite{deepseek2024v3} (72.2\%), despite being based on the same 671B parameter architecture. Similarly, Claude 3.7 Sonnet with a thinking budget of 1024 tokens (83.6\%) clearly outperformed the base Claude 3.7 Sonnet (80.4\%).
When compared with total cost of running the benchmark, we see that more expensive models tend to outperform less expensive models, either due to a larger underlying model or using more reasoning tokens in the output.

\subsection{Transferability of Reasoning Skills}
\begin{figure*}[th]
  \centering
  \begin{subfigure}{.32\linewidth}
    \includegraphics[width=\linewidth]{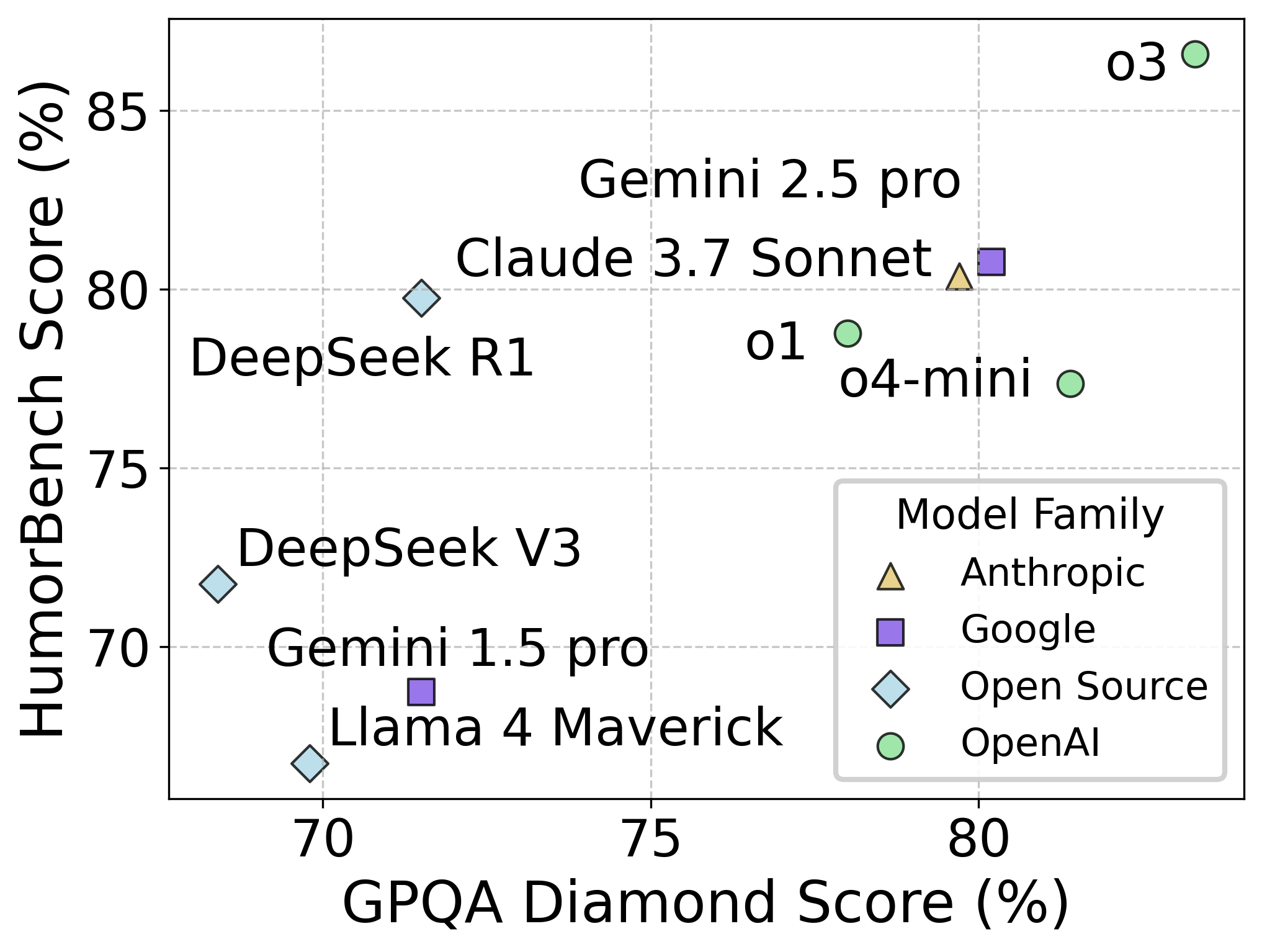}
    \label{fig:gpqa}
  \end{subfigure}
  \begin{subfigure}{.32\linewidth}
    \includegraphics[width=\linewidth]{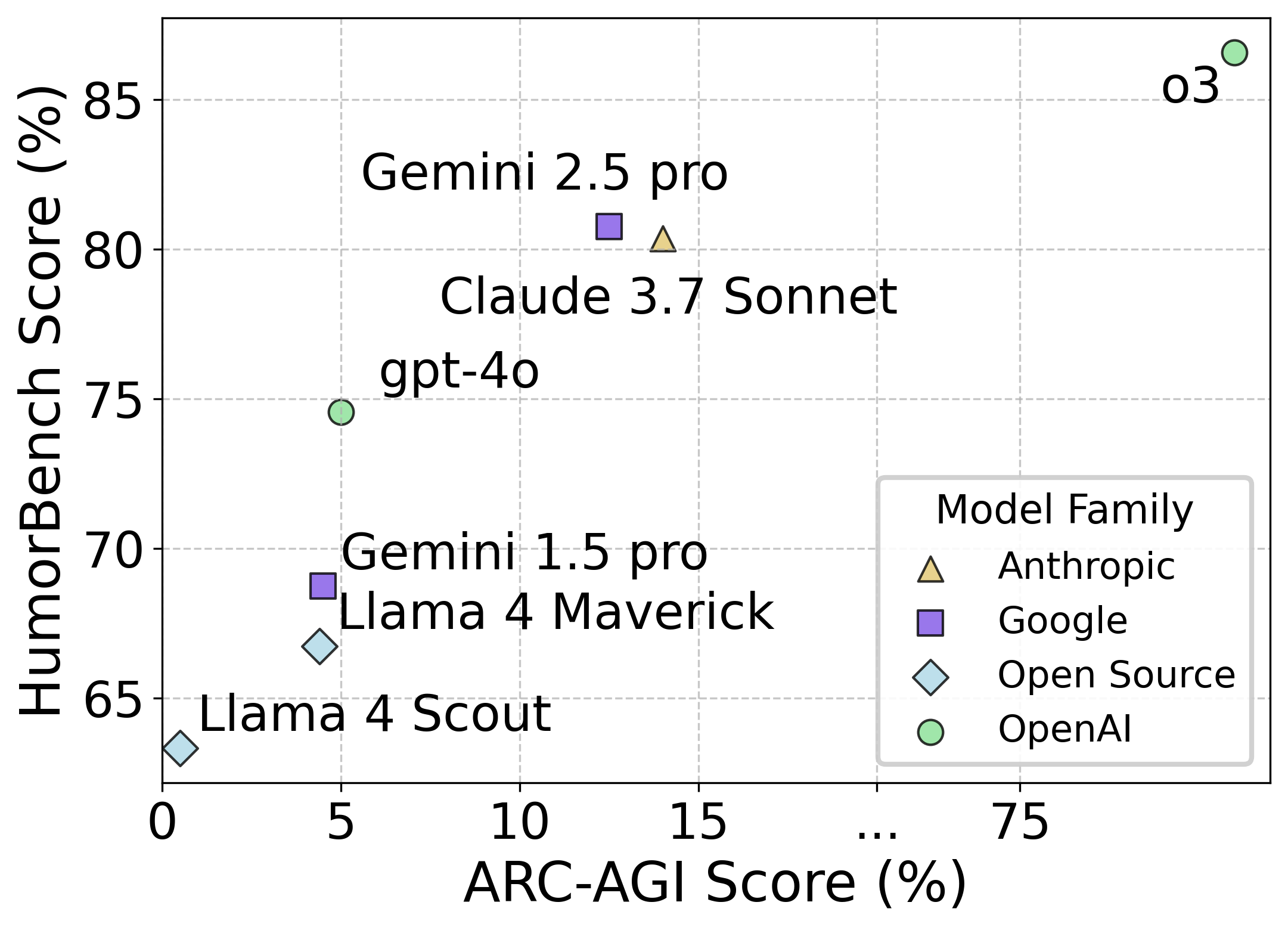}
    \label{fig:arc}
  \end{subfigure}
  \begin{subfigure}{.32\linewidth}      
    \centering                          
    \includegraphics[width=\linewidth]{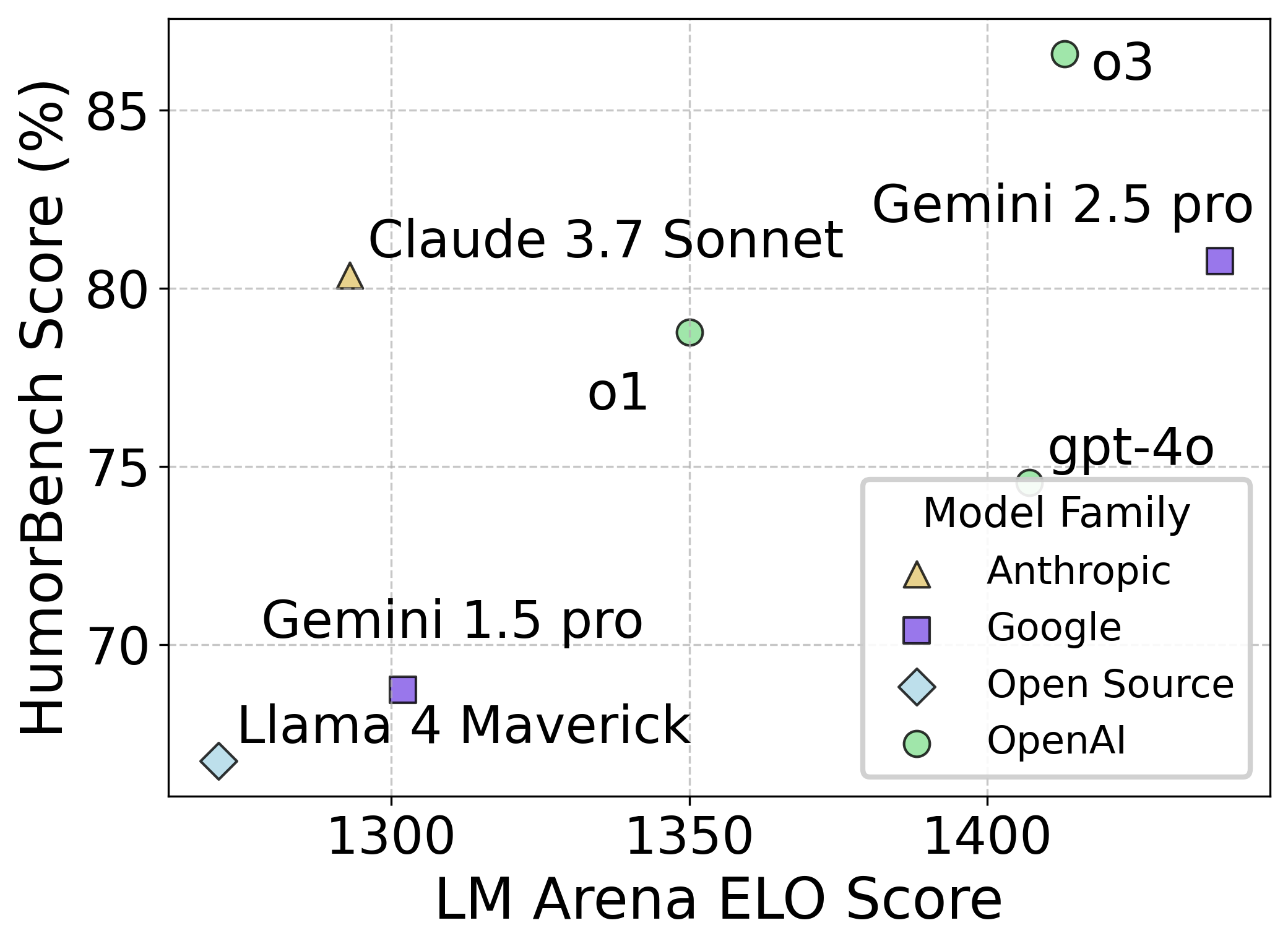}
    \label{fig:lmarena}
  \end{subfigure}

  \caption{HumorBench performance compared to several common benchmarks. We see positive correlation with GPQA, ARC-AGI, and LMArena. In particular, ranking compared to ARC-AGI is nearly identical to that of HumorBench, indicating a strong reasoning component to the HumorBench task.
  }
  \label{fig:transfer-2x2}
\end{figure*}

\begin{table}[ht!]
\centering
\small
\begin{tabular}{lcc}
\toprule
\textbf{Benchmark} & \textbf{Corr.} & \textbf{p-value} \\
\midrule
GPQA Diamond & 0.736* & 0.024 \\
ARC-AGI (with o-series) & 0.650 & 0.058 \\
ARC-AGI (w/o o-series) & 0.943** & 0.005 \\
LM Arena ELO & 0.714 & 0.071 \\
\bottomrule
\end{tabular}
\vspace{0.3em}
\caption{Spearman's rank correlations between HumorBench and other benchmarks. Asterisks indicate significance: *$p<0.05$, **$p<0.01$. ARC-AGI correlation shown separately for results with and without o-series models, which were fine-tuned for ARC-AGI}
\label{tab:benchmark-corr}
\end{table}
To gauge how well other model skills transfer to humor comprehension, we correlate HumorBench accuracy with three widely used LLM benchmarks: GPQA-Diamond \cite{rein2023gpqa}, ARC-AGI \cite{chollet2025arc}, and LM Arena ELO \cite{chiang2024arena}. HumorBench scores are positively associated with all three (see table \ref{tab:benchmark-corr}). In particular, after removing o-series models (whose reported scores come from ARC-tuned variants) the correlation with ARC-AGI rises to $\rho = 0.943$ ($p = 0.005$), underscoring the shared reasoning demands of the two tasks. The LM Arena correlation ($\rho = 0.714$) is solid but notably lower. Overall, this also suggests that LLM progress on STEM domains translates to Non-STEM reasoning as well. 

\subsubsection*{STEM-only reasoning improves HumorBench}
\begin{figure}[H]
  \centering
  \begin{subfigure}{.49\linewidth}
    \includegraphics[width=\linewidth]{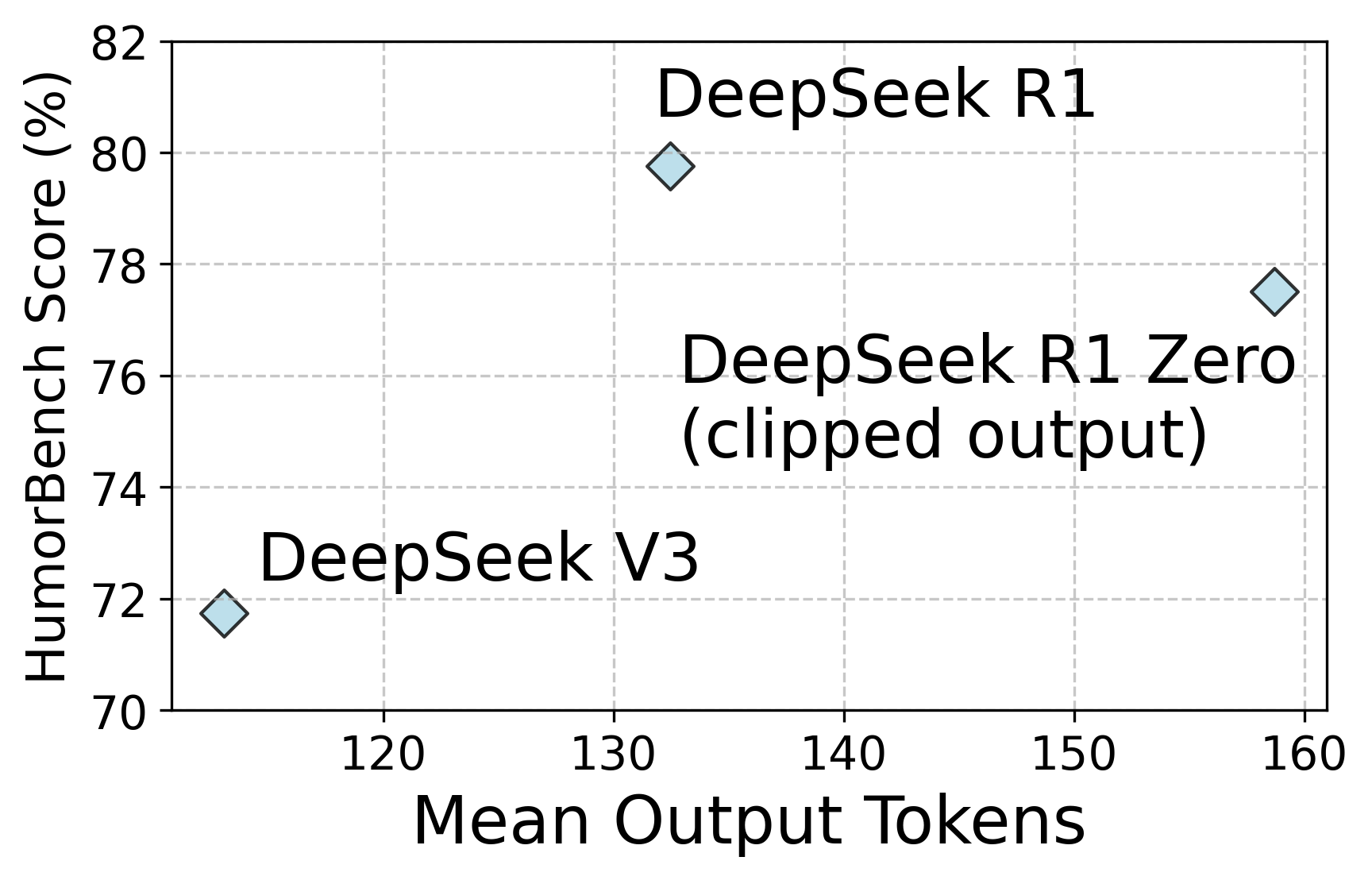}
    \label{fig:gpqa}
  \end{subfigure}\hfill
  \begin{subfigure}{.49\linewidth}
    \includegraphics[width=\linewidth]{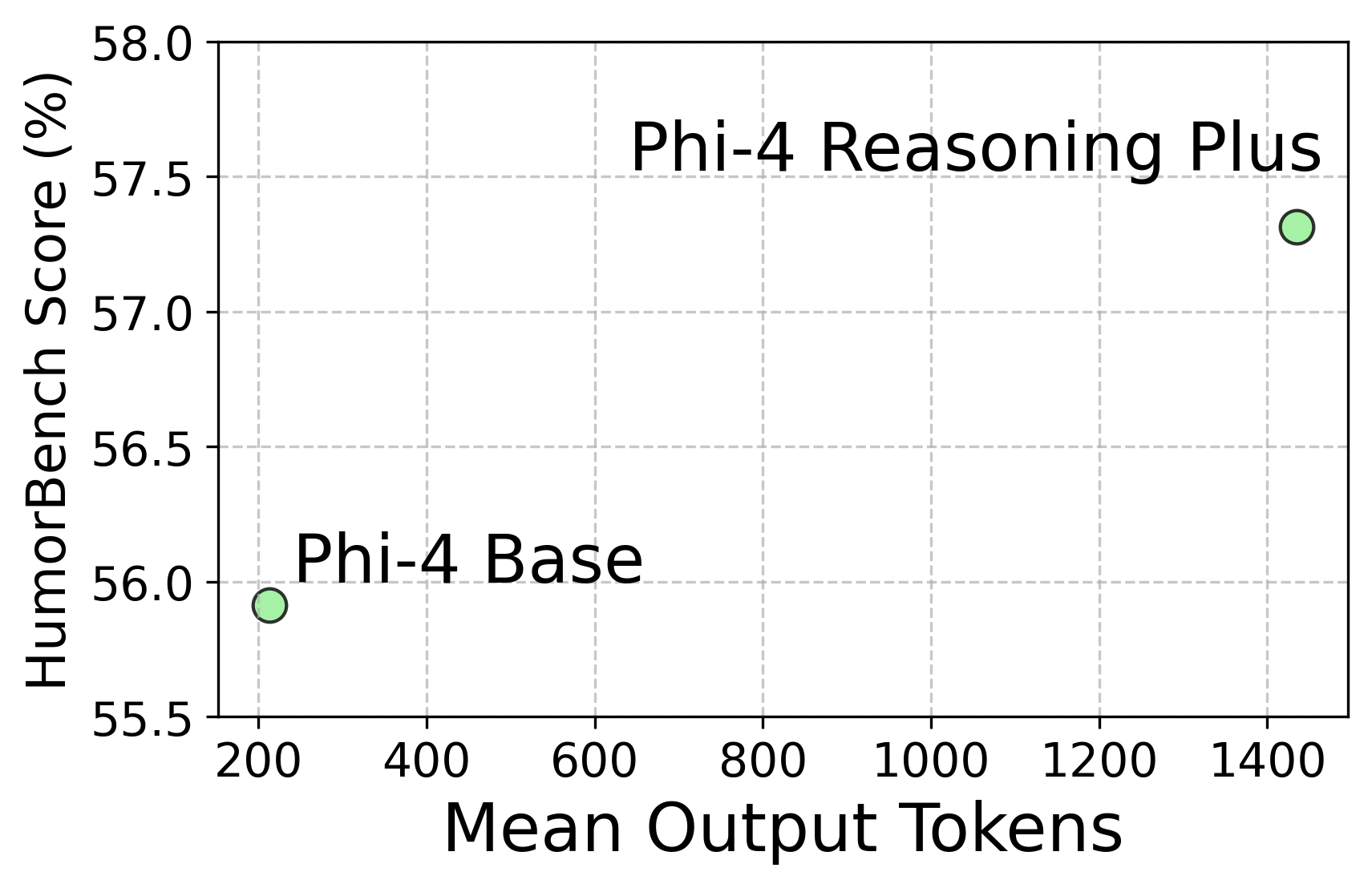}
    \label{fig:arc}
  \end{subfigure}
  \caption{Deepseek R1 Zero and Phi-4 Reasoning Plus, both exclusively reasoning-trained on STEM tasks, outperform their base versions on HumorBench}
  \label{fig:reasoning_r1}
\end{figure}

Our comparison between reasoning models trained on STEM tasks via Reinforcement Learning and their base counterparts yielded particularly revealing results. As illustrated in Figure~\ref{fig:reasoning_r1}, R1-Zero, which developed reasoning capabilities exclusively through self-play on STEM problems, demonstrated significant improvements over its base V3 model. Remarkably, it performed nearly on par with DeepSeek R1, despite the latter being trained on non-STEM data such as reading comprehension. Similarly, Phi-4 Reasoning Plus exhibited superior performance compared to its base model (Figure~\ref{fig:reasoning_r1}), although its training was limited to math and coding data~\citep{phi4reasoning2025}. These findings suggest that abstract reasoning capabilities are transferable to humor comprehension, indicating that the reasoning skills required for STEM domains may be fundamentally similar to those needed for understanding humor.

We also note that both R1-Zero and Phi-4 Reasoning Plus include their reasoning traces in their final outputs. Therefore, we evaluated their performances using the length control measure described above to ensure fair comparison across models.
\begin{figure*}[th]
  \centering
  \begin{subfigure}{.32\linewidth}
    \includegraphics[width=\linewidth]{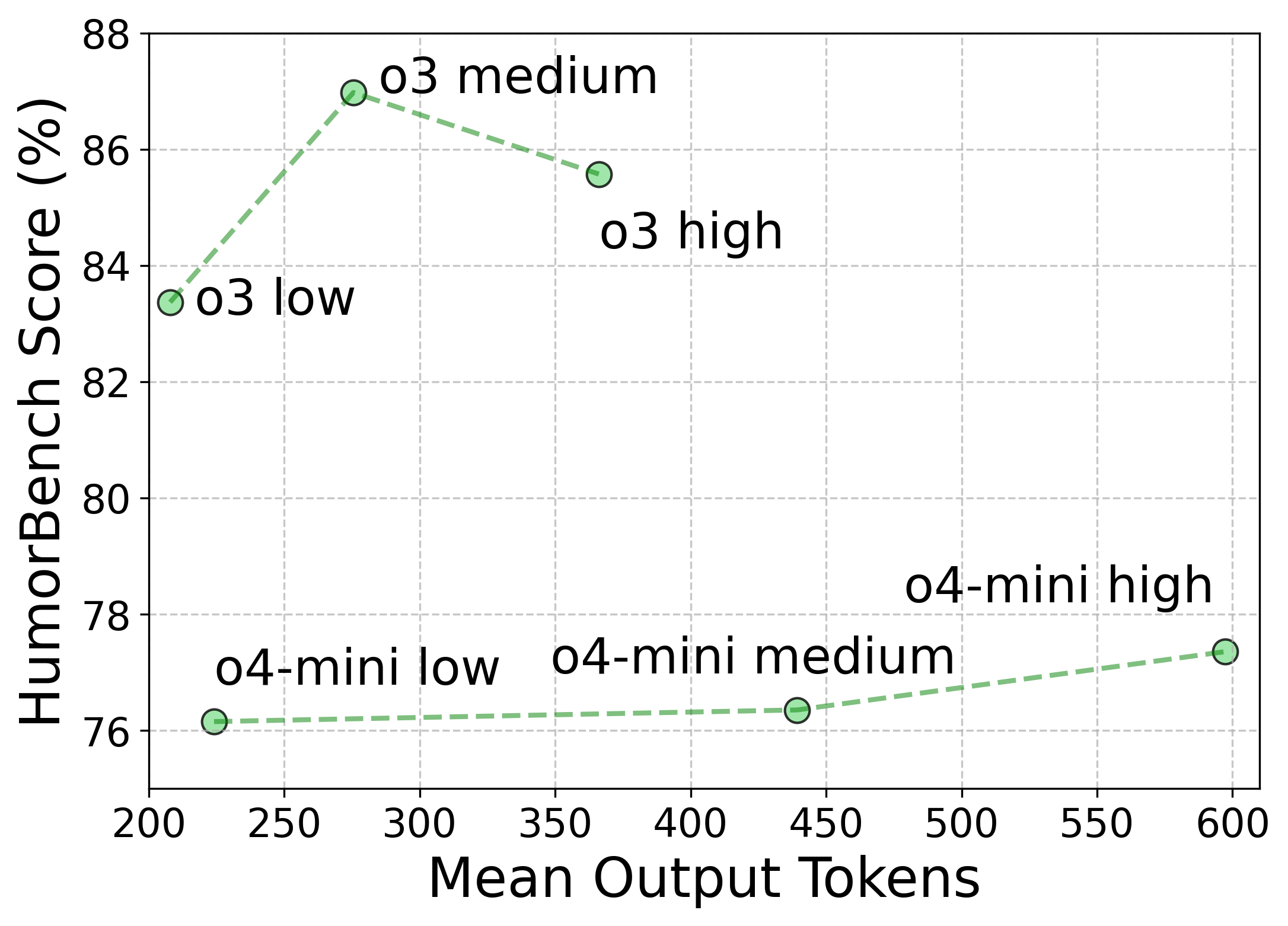}
    \caption{OpenAI o- series models, varying "reasoning effort"}
    \label{fig:gpqa}
  \end{subfigure}
  \begin{subfigure}{.32\linewidth}
    \includegraphics[width=\linewidth]{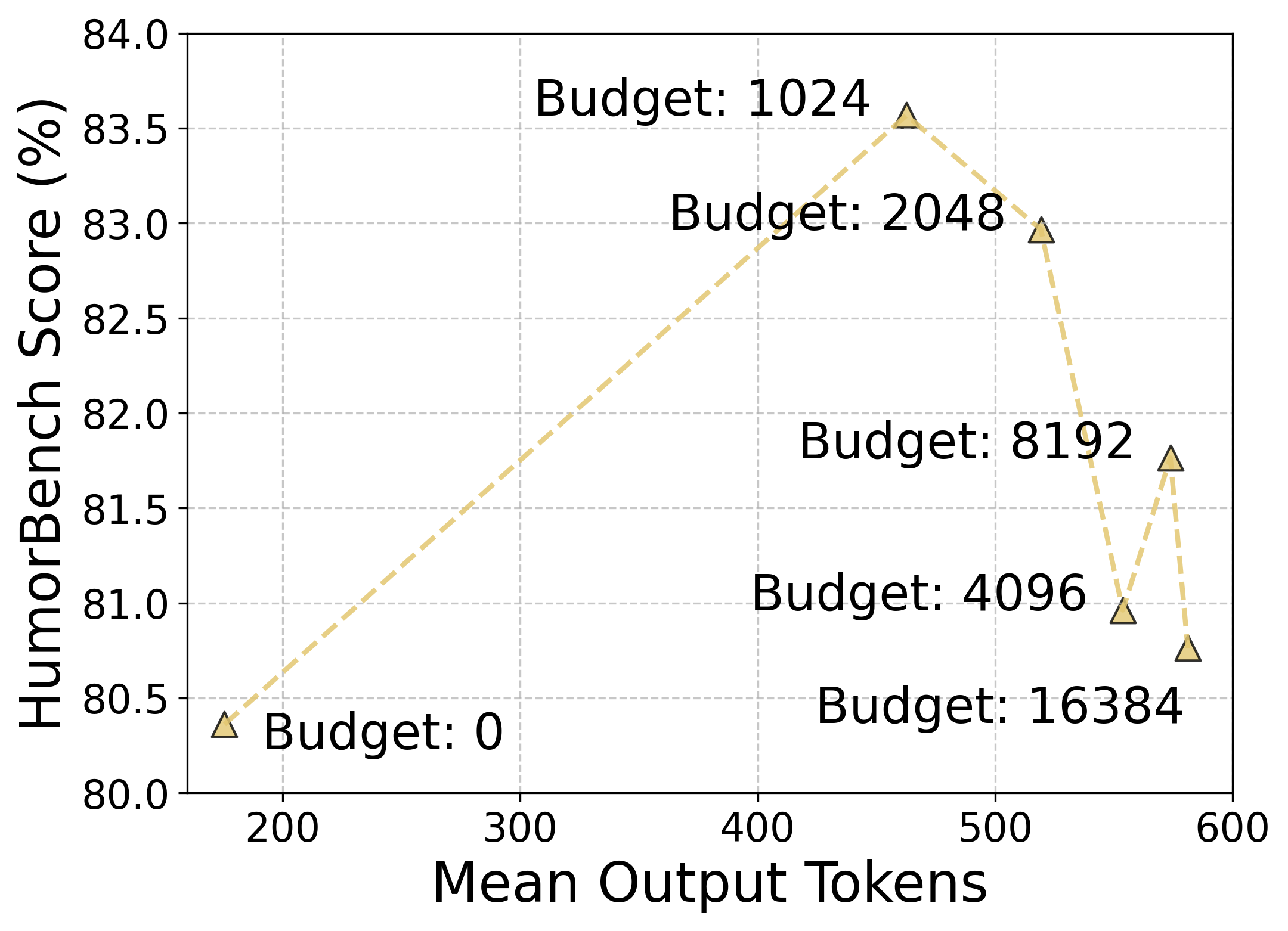}
    \caption{Claude 3.7 Sonnet performance, varying "thinking budget"}
    \label{fig:arc}
  \end{subfigure}
  \begin{subfigure}{.32\linewidth}
    \includegraphics[width=\linewidth]{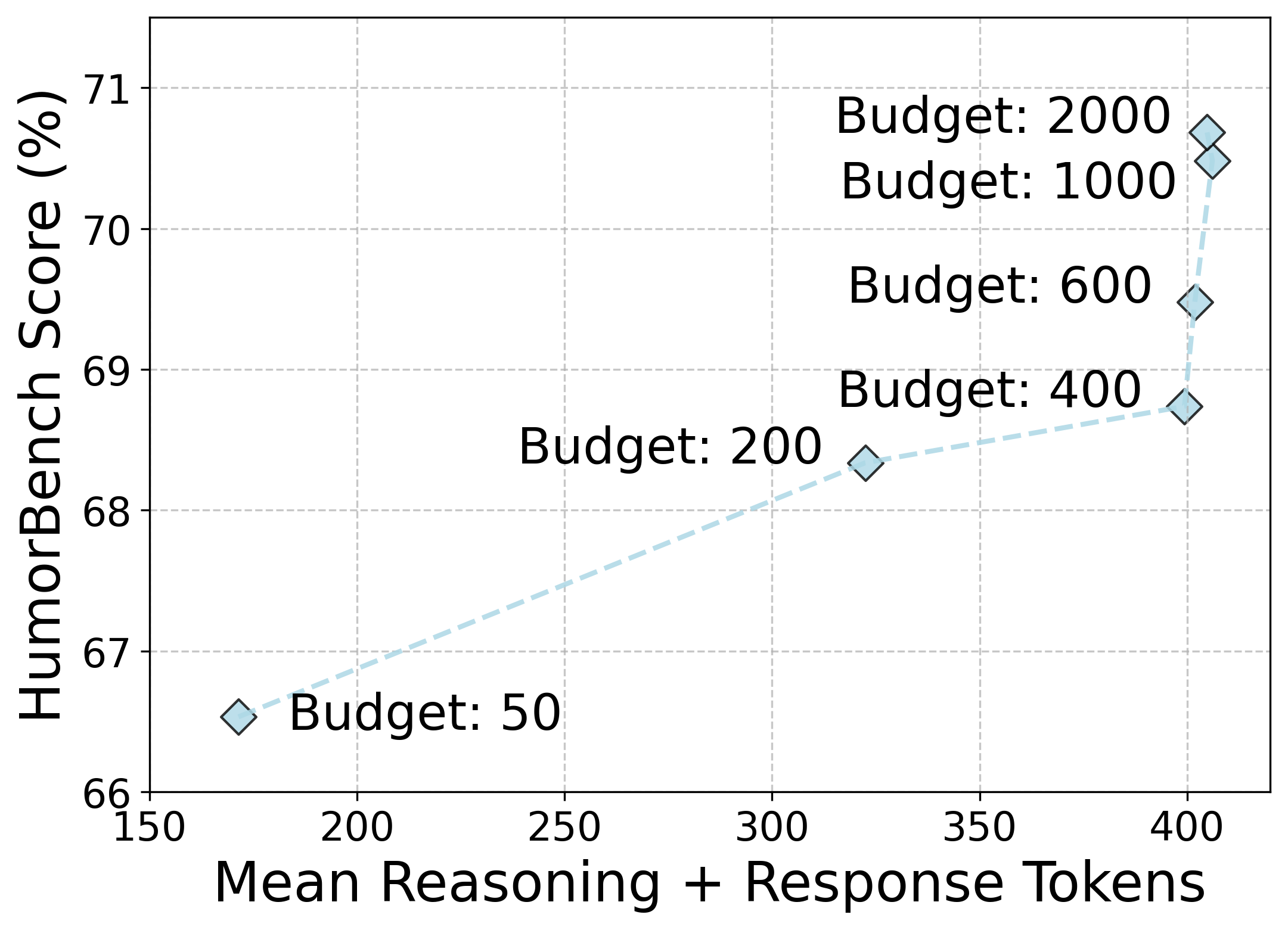}
    \caption{Qwen Plus performance, varying "thinking budget"}
    \label{fig:gpqa}
  \end{subfigure}\hfill

  \caption{HumorBench test-time compute experiments. Note, "mean output tokens" includes both reasoning and final response tokens}
  \label{fig:test_time}
\vspace{-\intextsep}
  
\end{figure*}

\subsection{Test-Time Scaling}

As seen in figure ~\ref{fig:test_time}, while including \textit{some} reasoning clearly helped model performance, the effect of continuing to increase test-time compute varied significantly between models. 
For Qwen plus and the o- series models, increasing the reasoning parameter (reasoning budget and "effort", respectively) generally improved performance. 
However, for Claude 3.7 Sonnet, increasing the thinking budget beyond the minimum 1024 tokens clearly \textit{hurt} performance. 

\begin{figure}[t]
    \centering
    \includegraphics[width=.9\linewidth]{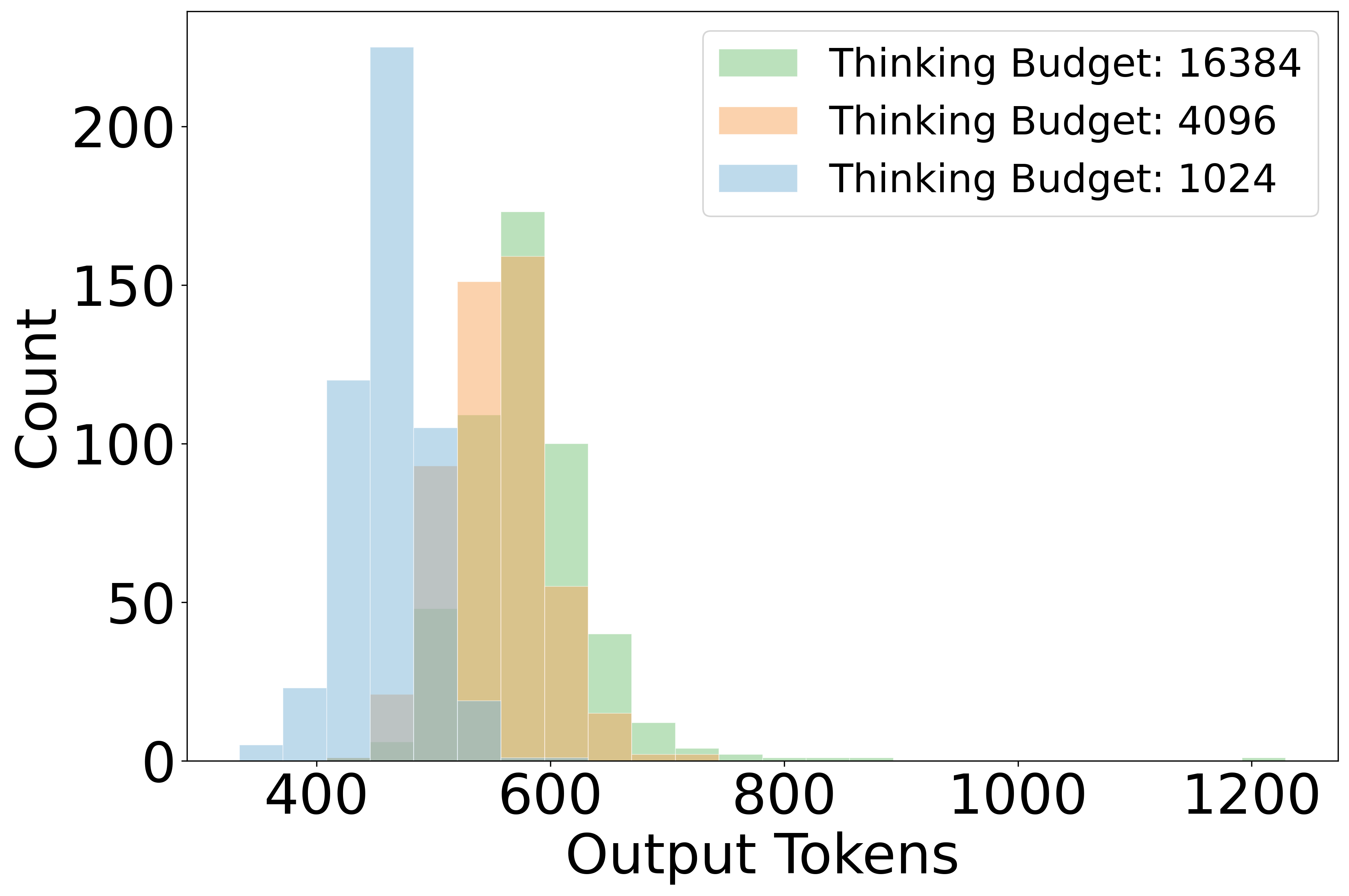}
    \caption{Token usage with different `thinking budget` parameters on Claude 3.7 sonnet. Most completions (in the hundreds of tokens) were far below their budgets.}
    \label{fig:claude_token_overlay}
\end{figure}

A closer look at the reasoning trace lengths highlights that for most captions, the models did not fully exhaust their reasoning budget (see ~\ref{fig:claude_token_overlay}).
Qwen Plus, for example, rarely used more than 400 tokens, even when budgeted 2000, a trend we saw for all test-time experiments. This suggests the LLMs are providing final answers based on completed thinking traces, which makes the inverse test-time scaling effect more puzzling. While a few studies have looked related problems~\citep{su2025between,shi2023large,yang2025towards,mckenzie2023inverse}, we defer thorough investigation of our observation to future work.

\subsection{Analysis of HumorBench Hard Subset}

\begin{figure*}[th]
  \centering
  \setlength{\tabcolsep}{4pt}
  \renewcommand{\arraystretch}{0}  
  \begin{tabular}{ccc}

  \subcaptionbox{%
    \footnotesize\textbf{Caption:} ``How about some help carrying the groceries?''\\
    \textbf{Element:} To the sharks, the person is the groceries%
  }[.32\linewidth]{%
    \includegraphics[height=0.18\textheight]{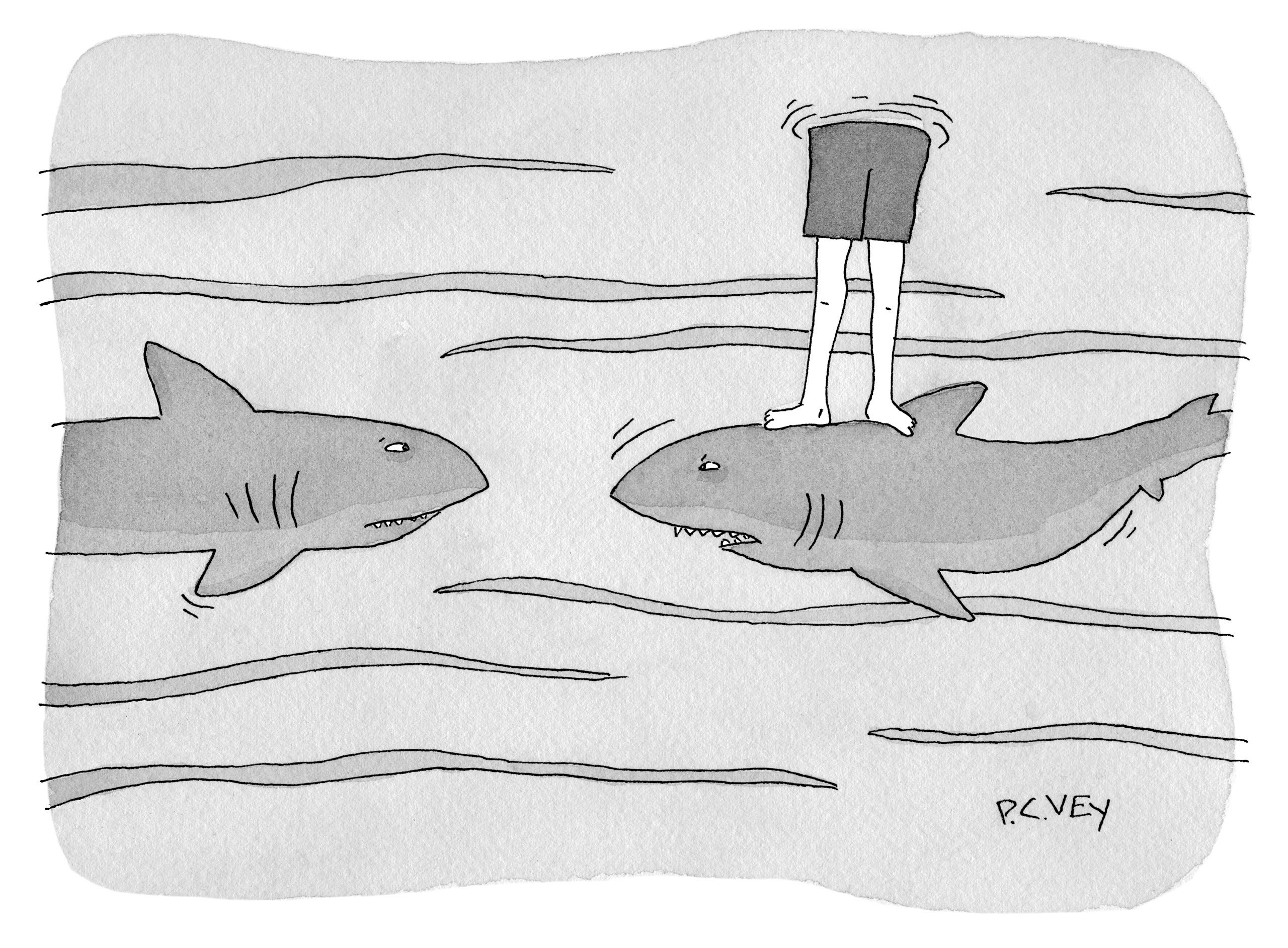}%
  } &

  \subcaptionbox{%
    \footnotesize\textbf{Caption:} ``I love his bedtime routine''\\
    \textbf{Element:} This is an explicit play on the dual meaning of the word “routine"", which could be either a child's “bedtime routine” or a comedian's “stand-up routine”%
  }[.32\linewidth]{%
    \includegraphics[height=0.18\textheight]{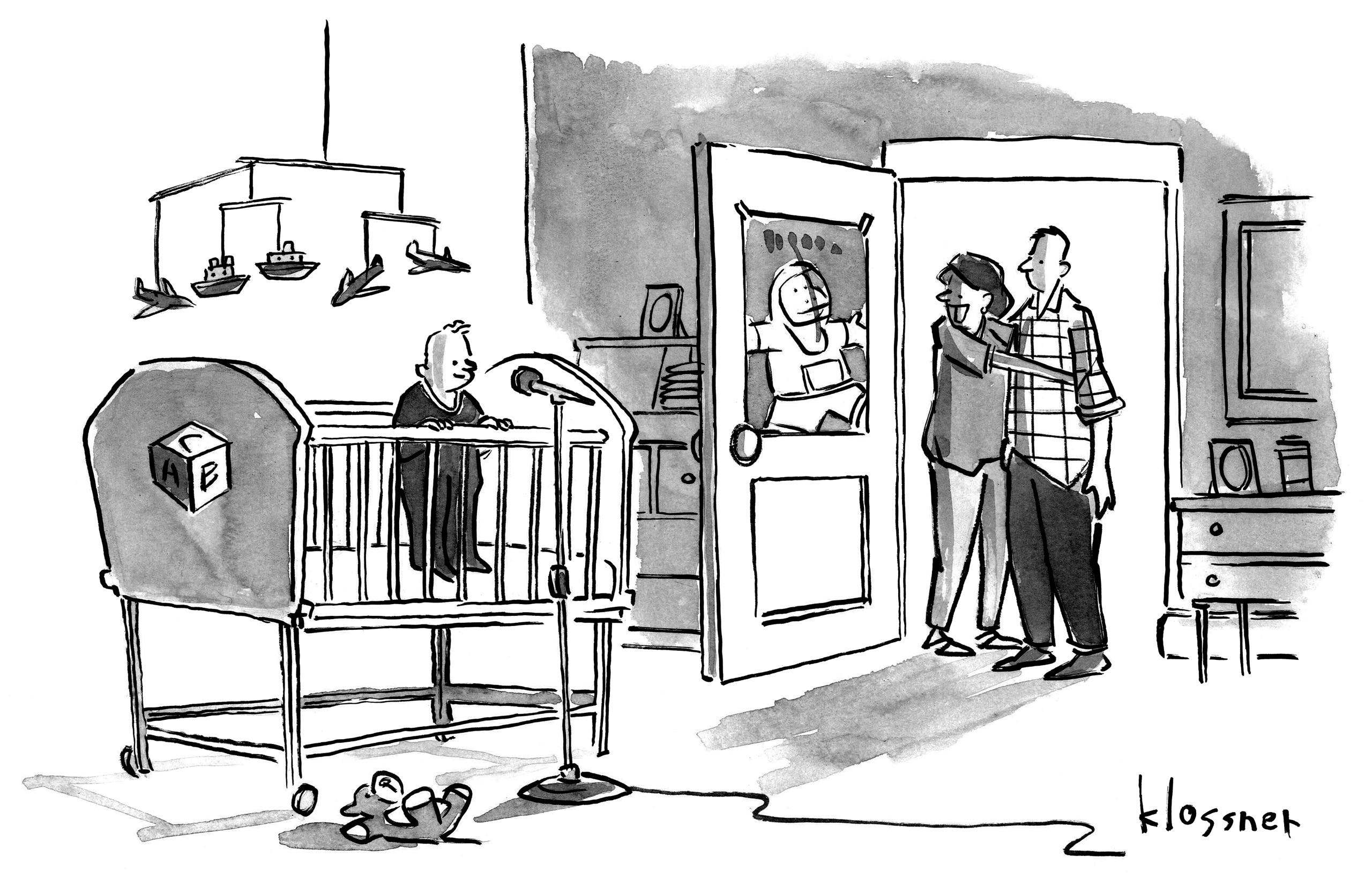}%
  } &

  \subcaptionbox{%
    \footnotesize\textbf{Caption:} ``I don't know how to tell him it isn't his.''\\
    \textbf{Element:} It should be obvious to him that the egg isn't his specifically because he's human, and humans don’t lay eggs%
  }[.32\linewidth]{%
    \includegraphics[height=0.18\textheight]{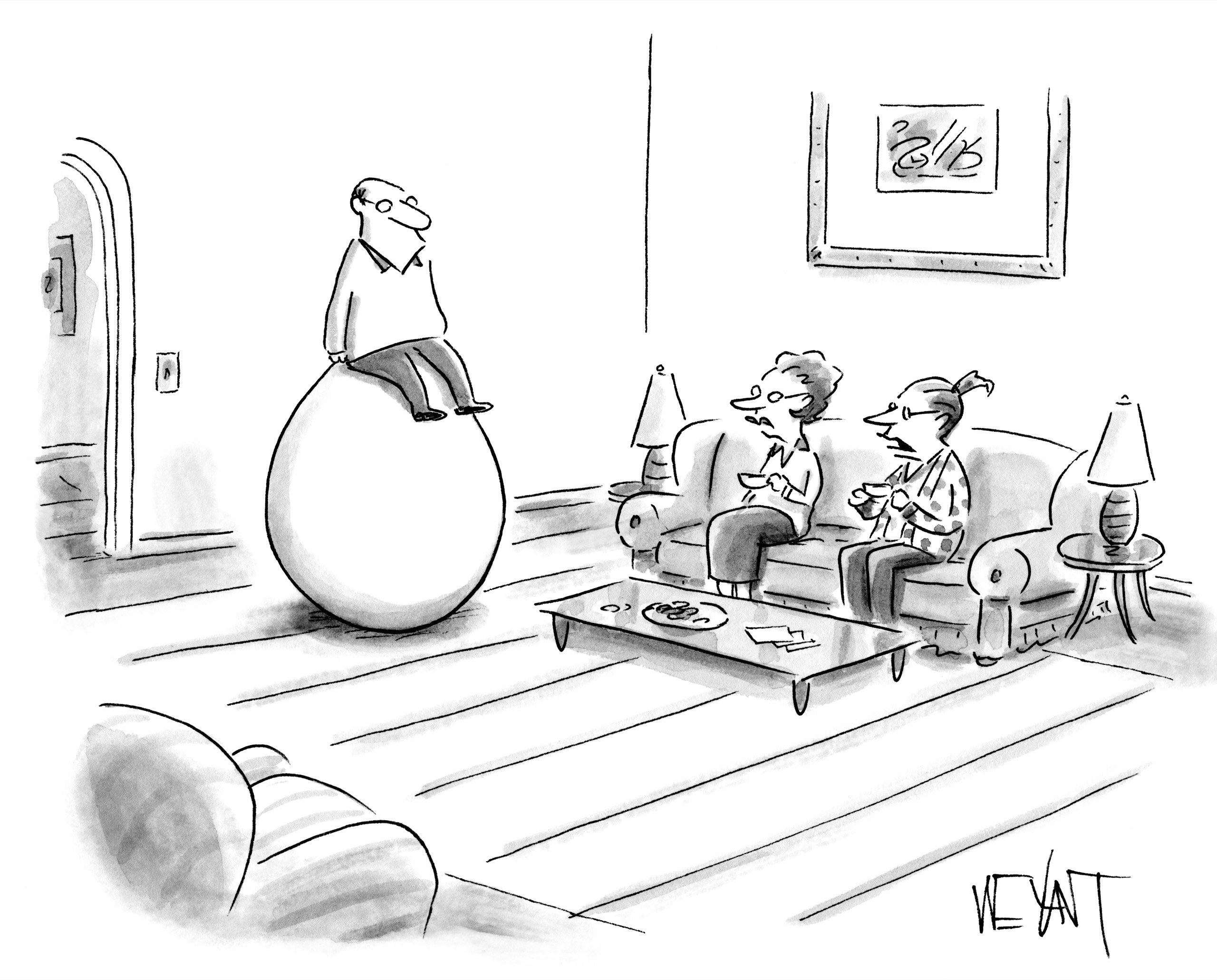}%
  } \\[6pt]

  \subcaptionbox{%
    \footnotesize\textbf{Caption:} ``This suit looked way better in the store.''\\
    \textbf{Element:} This is a play on words on "suit," as in a formal suit or a diving suit.%
  }[.32\linewidth]{%
    \includegraphics[height=0.18\textheight]{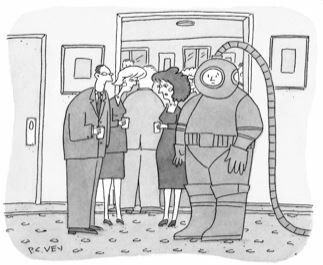}%
  } &

  \subcaptionbox{%
    \footnotesize\textbf{Caption:} ``It's painful, but I couldn't stand another chorus of ‘Take Me to the River.’''\\
    \textbf{Element:} This references Big Mouth Billy Bass, a classic novelty prop of a singing fish, singing "Take Me to the River"%
  }[.32\linewidth]{%
    \includegraphics[height=0.18\textheight]{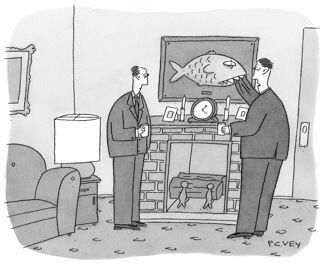}%
  } &

  \subcaptionbox{%
    \footnotesize\textbf{Caption:} ``How much did you spend at Macy’s this year?''\\
    \textbf{Element:} This implies the enormous person is a balloon for the Macy's day parade%
  }[.32\linewidth]{%
    \includegraphics[height=0.18\textheight]{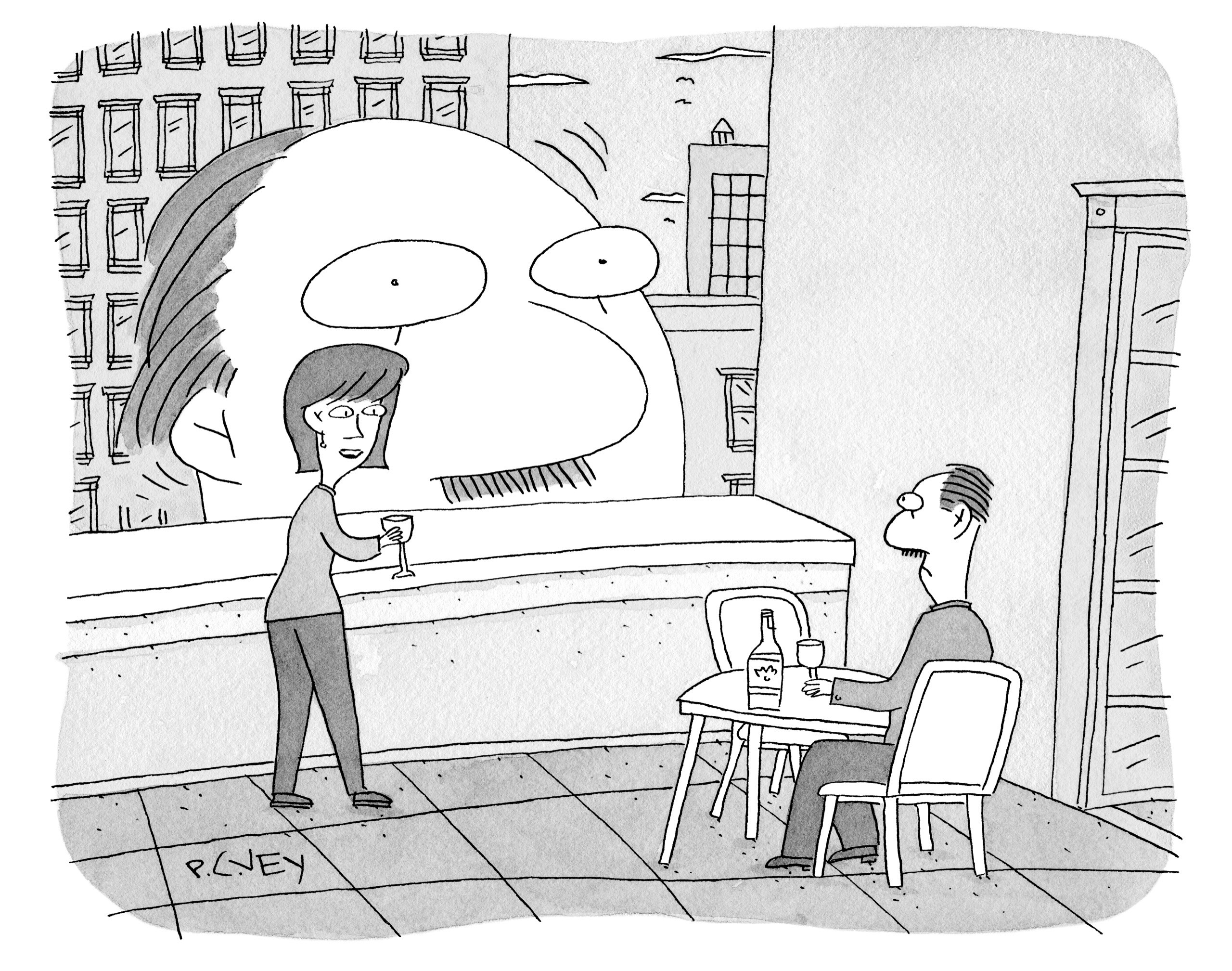}%
  }

  \end{tabular}

  \caption{Examples of elements in the \textsc{HumorBench} hard subset.  
           See Table~\ref{tab:hard-examples} for full descriptions.}
  \label{fig:hardset}
\end{figure*}

While frontier models like o3 demonstrate impressive performance on HumorBench, certain elements are persistently challenging for all models. To better understand the specific types of humor that remain challenging, we conducted a targeted analysis on the 100 unique elements that were most often missed during the benchmarking, which we call HumorBench Hard. Cartoons in this subset range from pass rates of 60\% (6 in 10 models get correct) to 0\% (No model gets correct).
See ~\ref{tab:hard-subset} for examples of the hard subset.

To get a more granular view of the elements that constitute the hard subset, we analyzed three predefined humor categories: \textit{wordplay}, \textit{cultural references}, and \textit{toxic or shocking} humor elements. These categories were annotated by an LLM categorization pipeline built on o3, which individually categorized each element as in or out of each category. Most elements did not fit into these categories, while some fit into multiple. Our analysis examined the relative representation of each category within the hard subset compared to their representation in the overall HumorBench dataset.

\begin{table}[t]
\centering
\small
\begin{tabular}{lccc}
\toprule
Humor Category & Entire Set & Hard Subset & Diff (\%)\\
\midrule
Wordplay & 24.4\% & 19.0\% & -5.4\\
Cultural Reference & 19.0\% & 17.7\% & -1.3\\
Toxic or Shocking & 25.7\% & 26.6\% & +0.9\\
\bottomrule
\end{tabular}
\caption{Representation of humor categories in the full dataset compared to the hard subset.}
\label{tab:hard-subset}
\end{table}

We summarize the main findings of this analysis in Table ~\ref{tab:hard-subset}. Overall, these relatively minor deviations indicate that humor category alone is not the primary determinant of difficulty for models. Challenging examples likely hinge on subtler factors, such as the implicit conceptual leaps required or the obscurity or references. We observed that wordplay was slightly \textit{under-represented} among the hard subset ($-5.4\%$), suggesting current LLMs handle puns or jokes that rely on linguistic manipulation somewhat better than other elements. \emph{Cultural references} and \emph{toxic or shocking} humor, meanwhile, were essentially evenly represented, indicating that these styles do not disproportionately increase difficulty. These nuanced insights encourage further qualitative investigation into the underlying reasons why particular humor instances remain difficult, even for state-of-the-art LLMs trained explicitly with reasoning capabilities.

Lastly, in Figure~\ref{fig:hardset}, we highlight a few concrete examples that are particularly challenging in the HumorBench hard subset. Detailed descriptions are also provided in Table~\ref{tab:hard-examples}. To our surprise, although several of them are quite intuitive and easy to recognize for humans, LLMs usually struggle on these examples. This represents a fundamental difference in reasoning abilities between humans and LLMs.


\section{Conclusion and Future Work}

In this work, we introduced \textbf{HumorBench}, the first large-scale evaluation that isolates \emph{humor comprehension}, as opposed to subjective funniness, by grading model explanations against concise, expert-annotated \emph{objective elements}. Our experiments with more than a dozen frontier and open-source LLMs revealed that (i) progress on STEM reasoning benchmarks translates strongly to non-STEM humor reasoning, (ii) specialized "reasoning" variants consistently outperform base models even when they were trained only on STEM corpora, and (iii) test-time compute helps, but only up to the point where the relevant background knowledge is actually present in the model. Together, these findings position HumorBench as a sensitive probe of higher-level reasoning that remains comfortably unsolved: the best model still misses over 40\% of elements in our hard subset.

Looking forward, there are a number of promising avenues for extending this work. For example, a natural next step is to develop a \emph{multimodal} version of HumorBench by reintroducing the original cartoon images. This would allow evaluation of both visual recognition and reasoning, better reflecting the complete humor comprehension task. Alternatively, the use of HumorBench’s element rubrics as intermediate supervision signals offers an exciting opportunity to explore fine-tuning or reinforcement learning. Finally, improving the reliability of LLM-as-judge evaluation remains an important open challenge, particularly for creative tasks with high output variability.

We hope HumorBench spurs progress on reasoning that bridges the gap between logical deduction and nuanced human culture, and serves as a springboard for genuinely funny, culturally aware AI systems.

\newpage
\section*{Limitations}

HumorBench focuses on textual explanations derived from detailed image descriptions. While this controls for pure vision problems, it inevitably removes part of the challenge: recognizing visual cues. The autograder, despite a 92\% agreement with humans, is mildly lenient, so reported scores should be seen as upper bounds. Additionally, while extensive validation and refinement were applied, some element annotations may still contain errors or reflect subjective interpretations that slipped through, introducing noise into model evaluation. Finally, the dataset size (499 unique elements) limits the statistical resolution of fine-grained analyses.



\bibliography{custom}

\onecolumn
\appendix

\section{Main Benchmark Prompt}
\label{sec:Explainer_Prompt}
\bigskip                                
\begin{center}
    \begin{minipage}{0.93\linewidth}    
      \ttfamily\small                   

You are a humor expert extraordinaire, judging the New Yorker Cartoon Caption Contest. Your current task is to help us understand the humor in various submitted captions. Given a cartoon description and a caption submission, explain (in less than 200 words) *what* the joke is, focusing on the material substance of the joke.
STRICTLY use the format: <explanation>explanation goes here</explanation>

Cartoon description: {description}
Caption: {caption}
    \end{minipage}
\end{center}
\bigskip  

\section{Autograder Prompt}
\label{sec:Autograder Prompt}
\bigskip                                
\begin{center}
    \begin{minipage}{0.93\linewidth}    
      \ttfamily\small                   

You will receive:
1. A short cartoon description
2. A winning funny caption
3. A student’s answer
4. A brief “anticipated answer point” that captures the crucial comedic device or element

Your job is to determine whether the student’s answer **explicitly covers** that “anticipated answer point.” 

- If the student’s answer captures or discusses the key comedic element (even if the wording is different), **PASS**.
- If the student’s answer **omits** or **contradicts** that key comedic element, **FAIL**.
- Do not penalize extra details or expansions. Synonyms or paraphrasing are acceptable if they convey the same comedic logic.
- Be mindful: if the anticipated answer point emphasizes something specific (e.g. a pun, wordplay, or ironic twist), check that the student’s answer includes it.

At the end of your evaluation, provide exactly two XML tags:
1. <reasoning>Short explanation of your thought process</reasoning>
2. <judgement>PASS or FAIL</judgement>

Do not include additional commentary or deviation from this format.

Cartoon description: \{description\}
Caption: \{caption\}
Student's answer: \{explanation\}
Anticipated answer point:\{anticipated point\}
\end{minipage}
\end{center}
\bigskip  

\section{Hard Example Details}
\label{sec:hard-examples}

\begin{table}[h!]
  \centering
  \footnotesize
  \begin{adjustbox}{max width=\textwidth}
      \begin{tabular}{lp{0.37\textwidth}p{0.13\textwidth}p{0.27\textwidth}c}
      \toprule
      \textbf{Cartoon ID} & \textbf{Description} & \textbf{Caption} & \textbf{Element(s)} & \textbf{Pass Rate (\%)} \\
      \midrule
      CC123065 & Inside a workshop like room, three elves in pointy hats sit at a long table with open laptop computers. The middle elf appears distressed and is speaking, while the two elves on either side look toward him. & ``It’s from Santa, and it goes way, way beyond jolly.'' & Frames Santa as a boss making an inappropriate advance on an employee. & 0 \\
      \addlinespace
      NYCC \#40 & The cartoon shows a woman in her underwear sitting up in bed, looking forward with a disgruntled expression. A large snow globe with a snowman inside is positioned next to her on the bed. The woman is speaking. & ``I think the Manhattan skyline is getting suspicious.'' & This implies that she is cheating on her partner, a snow globe of the Manhattan skyline, with the snowman. & 17 \\
      \addlinespace
      NYCC \#15 & In a restaurant, a man and a woman are sitting down to eat dressed in nice clothing. The man is leaning over the table with his hand on a glass looking at the woman with a soft smile. However, the man is bald and has a cartoonishly large forehead, with the outline of the woman visible on his forehead. The woman, sitting upright, is speaking. & ``Well, it's a lovely gesture, but I still think we should start seeing other people.'' & Implies that the image on his forehead is a tattoo, as getting a tattoo of your significant other is a common practice. & 20 \\
      \addlinespace
      NYCC \#669 & A baby leans over the side of a crib toward a microphone on a stand, as if ready to perform. The crib has letter blocks, and a mobile with boats and airplanes hangs above. A teddy bear lies on the floor. In the background, a couple stands in the doorway, looking at the baby with surprise. The woman is speaking with a smile. & ``I love his bedtime routine.'' & Play on the dual meaning of the word ``routine'': a child's bedtime routine vs. a comedian's stand up routine. & 33 \\
      \addlinespace
      NYCC \#665 & Two sharks are facing each other in the ocean. A person, visible only from the waist down, is standing on the back of one of the sharks. The sharks look bewildered; the carrying shark is speaking. & ``How about some help carrying the groceries?'' & To the sharks, the person is the groceries. & 40 \\
      \addlinespace
      NYCC \#687 & A woman holding a wine glass stands on a rooftop in a city, delighted, looking back at a man sitting at a table with a bottle and glass. Behind them, an enormous face peers over the building, resembling the man. The woman is speaking with a smile. & ``How much did you spend at Macy’s this year?'' & Implies the enormous person is a parade balloon for the Macy’s Thanksgiving Day Parade. & 40 \\
      \addlinespace
      NYCC \#686 & In a living room, a bald man is sitting on a giant egg, looking content. Two older women sipping tea, seated on a couch, are staring at him. The room has a coffee table, lamps, and a framed picture on the wall. One woman is speaking. & ``I don't know how to tell him it's not his.'' & It should be obvious to him that the egg isn't his because humans don’t lay eggs. & 27 \\
      \addlinespace
      NYCC \#61 & A doctor wearing a head mirror stands behind a desk in a typical office. A giant hand is reaching through the doorway, palm up. The doctor is leaning over to check the enormous hand's pulse. & ``I don't know why you're so jolly---your cholesterol is through the roof.'' & Wordplay: ``through the roof'' both as extremely elevated levels and literally breaking through the roof. & 20 \\
      \addlinespace
      \bottomrule
    \end{tabular}
  \end{adjustbox}
  \caption{Representative examples from the hard subset where a majority of evaluated LLMs failed to identify all required humor elements.}
  \label{tab:hard-examples}
\end{table}

\section{Model performance on HumorBench Hard}
\begin{table}[t]
  \centering
  \footnotesize
  \begin{tabular}{lc}
    \toprule
    \textbf{Model} & \textbf{Accuracy (\%)} \\
    \midrule
    o3                                   & 59.85 \\
    Claude 3.7 Sonnet                    & 54.27 \\
    gemini-2.5-pro-preview-03-25         & 52.44 \\
    deepseek/deepseek-r1-zero            & 51.22 \\
    deepseek-ai/DeepSeek-R1              & 51.22 \\
    o1                                   & 50.00 \\
    o4-mini                              & 46.34 \\
    Grok 3                               & 45.12 \\
    gpt-4o                               & 42.68 \\
    deepseek-ai/DeepSeek-V3              & 39.63 \\
    meta-llama/Llama-4-Maverick-17B-128E & 35.98 \\
    gemini-1.5-pro                       & 35.37 \\
    o3-mini                              & 32.93 \\
    meta-llama/Llama-4-Scout-17B-16E     & 29.88 \\
    Qwen/Qwen2.5-72B-Instruct-Turbo      & 26.83 \\
    \bottomrule
  \end{tabular}
  \caption{Accuracy on the \textbf{HumorBench-Hard} subset (100 items).}
  \label{tab:hard-subset-model-perf}
\end{table}

\end{document}